\documentclass[final, times, 12pt]{elsarticle}
\usepackage{amsmath,amsfonts,amssymb,amsthm}
\usepackage{booktabs}
\usepackage[colorlinks=true, allcolors=blue]{hyperref}
\usepackage{algorithmic}
\usepackage{algorithm}
\usepackage{setspace}
\usepackage{array}
\usepackage[caption=false,font=normalsize,labelfont=sf,textfont=sf]{subfig}
\usepackage{textcomp}
\usepackage{stfloats}
\usepackage{url}
\usepackage{tikz}
\usepackage{verbatim}
\usepackage{lettrine}
\usepackage{graphicx}
\newcommand{\elsPARstart}[2]{%
    \lettrine[lines=2, lhang=0.1, findent=0.1em, nindent=0em]{\textsc{#1}}{#2}%
}

\newtheorem{lemma}{lemma}
\newcommand{\GeoTop}{\textsc{GeoTop}}

\newcommand{\PD}{\text{PD}}
\newcommand{\bottleneck}{d_B}
\newcommand{\wasserstein}[1]{d_W^{#1}}

\begin{document}

\begin{frontmatter}

\title{GeoTop: Advancing Image Classification with Geometric-Topological Analysis}

\author[1]{Mariem Abaach}
\author[2,3]{Ian Morilla\corref{mycorrespondingauthor}}

\address[1]{MAP5 UMR CNRS 8145, Universit\'e de Paris,
            45 rue des Saints-P\`eres,
            75006 Paris, France}

\address[2]{Universit\'e Sorbonne Paris Nord, LAGA, CNRS, UMR 7539, 
            Laboratoire d'excellence Infibrex,
            99 avenue Jean Baptiste cl\'ement,
            F-93430 Villetaneuse, Paris, France}

\address[3]{Instituto de Hortofruticultura Subtropical y Mediterr\'anea La Mayora (IHSM), 
            Universidad de M\'alaga-Consejo Superior de Investigaciones Cient\'ificas,
            Boulevar Louis Pasteur, 49,
            29010 M\'alaga, Spain}

\cortext[mycorrespondingauthor]{Corresponding author: morilla@math.univ-paris13.fr or ian.morilla@ihsm.uma-csic.es}

\begin{abstract}
A fundamental challenge in diagnostic imaging is the phenomenon of topological equivalence, where benign and malignant structures share global topology but differ in critical geometric detail, leading to diagnostic errors in both conventional and deep learning models. We introduce GeoTop, a mathematically principled framework that unifies Topological Data Analysis (TDA) and Lipschitz-Killing Curvatures (LKCs) to resolve this ambiguity. Unlike hybrid deep learning approaches, GeoTop provides intrinsic interpretability by fusing the capacity of persistent homology to identify robust topological signatures with the precision of LKCs in quantifying local geometric features such as boundary complexity and surface regularity.

The framework's clinical utility is demonstrated through its application to skin lesion classification, where it achieves a consistent accuracy improvement of 3.6\% and reduces false positives and negatives by 15--18\% compared to conventional single-modality methods. Crucially, GeoTop directly addresses the problem of topological equivalence by incorporating geometric differentiators, providing both theoretical guarantees (via a formal lemma) and empirical validation via controlled benchmarks. Beyond its predictive performance, GeoTop offers inherent mathematical interpretability through persistence diagrams and curvature-based descriptors, computational efficiency for large datasets (processing \(224\times224\) pixel images in \(\leq 0.5\)\ s), and demonstrated generalisability to molecular-level data. By unifying topological invariance with geometric sensitivity, GeoTop provides a principled, interpretable solution for advanced shape discrimination in diagnostic imaging.

\end{abstract}


\end{frontmatter}

\section{Introduction}
\elsPARstart{M}{odern} medical imaging generates complex, high-dimensional data that captures multi-scale structural information critical for disease diagnosis and treatment monitoring \cite{Cancerdata,RANJBAR2017223,Singh2023}. While essential for clinical decision-making, this data presents significant analytical challenges: conventional methods, including deep learning approaches \cite{LITJENS201760,Wang2024}, often fail to simultaneously capture global topological patterns and local morphological details that are crucial for accurate diagnosis. These limitations are particularly critical in cancer diagnostics, where subtle topological and geometric features can distinguish malignant from benign lesions \cite{Vandaele2020,Somasundaram2021-bl}. Although advances in radiomics \cite{RANJBAR2017223} and spatially resolved transcriptomics \cite{Rahman2022,Marx2021,Hasin2017} have improved feature extraction, robust characterisation of pathological morphology remains challenging due to fundamental limitations in shape representation.

A fundamental challenge in diagnostic image analysis involves distinguishing pathological structures that are geometrically distinct yet topologically equivalent. Classical topological data analysis (TDA), while robust to noise and invariant under continuous deformations, is inherently limited by its homeomorphic invariance: persistence diagrams capture the birth and death of topological features but disregard crucial metric attributes such as curvature, boundary complexity, and surface regularity \cite{Carlsson2009,RobinsonT17}. This limitation becomes critical in clinical applications where topological equivalence masks diagnostically significant geometric variation--for instance, when differentiating between benign moles with smooth boundaries and malignant melanomas with irregular, infiltrating margins \cite{Carlsson_Vejdemo-Johansson_2021,doi:10.1073/pnas.1102826108,Singh2007}. Overcoming this limitation requires descriptors that maintain TDA's invariance properties while incorporating sensitivity to geometric structure.

While several approaches have attempted to enrich persistent homology with additional information, most extensions remain either data-driven augmentations (e.g., neural embeddings of persistence diagrams) or task-specific hybrids lacking a principled geometric foundation \cite{HoferKNU17,CarriereCO17}. Methods such as topological signatures, persistence images, or learned kernel embeddings improve discriminative capacity empirically but rarely address the fundamental source of ambiguity: the loss of metric information under homeomorphic invariance \cite{AdamsEKNPSCHMZ17,bubenik2015statisticaltopologicaldataanalysis,ReininghausHBK15}. Conversely, purely geometric descriptors, while sensitive to curvature and scale, fail to capture the hierarchical structure and stability that make TDA valuable for noisy medical images \cite{chazal2021introduction,bronstein2016geometric,Oudot15,Memoli11}.

We introduce \textbf{GeoTop}, a unified framework that bridges this gap by systematically integrating Lipschitz-Killing Curvatures (LKCs--which quantify intrinsic volumes across thresholds--with persistent homology. This integration yields a geometric-topological representation that resolves diagnostic equivalence classes invisible to topology alone. GeoTop derives features from both perspectives: it encodes global topology through persistent homology and local geometry via LKCs, unifying them within a consistent mathematical framework grounded in integral geometry. This fusion not only enhances classification accuracy but also provides interpretable descriptors, enabling the decomposition of discriminative signals into topological versus geometric contributions--a crucial property for clinical applications where understanding \textit{why} a model distinguishes malignant from benign cases is as important as \textit{how} it does so.

TDA and LKCs offer complementary solutions to these challenges. TDA, through persistent homology \cite{AdlerTDA,Hatcher,Singh2023}, identifies robust topological signatures that persist across scales, such as the connected components and voids characteristic of tumour microarchitecture \cite{Morilla2022,Rabadan2019}. Meanwhile, LKCs \cite{geotopo} quantify precise geometric properties including boundary complexity and surface regularity (Figure~\ref{fig:pipeline}). While effective individually--with applications ranging from skin cancer detection \cite{skinTDA} to radiomics \cite{Vandaele2022}--their combination remains under-explored despite the potential for synergistic performance gains in medical image analysis.

Our GeoTop framework addresses three critical gaps in diagnostic imaging: (1) \textit{The diagnostic equivalence problem}--benign and malignant structures may share identical topology but differ geometrically in ways that are clinically significant; (2) \textit{Multi-scale integration}--most methods analyse either global topology or local geometry, but not their interaction across spatial scales; and (3) \textit{Interpretability}--current deep learning approaches often operate as black boxes, lacking the mathematical transparency needed for clinical trust.

Through comprehensive validation on 3,297 clinically-annotated skin lesions (1,800 benign/1,497 malignant), we demonstrate that GeoTop:
\begin{itemize}
    \item Achieves consistent accuracy improvements of 3.6\% over single-modality methods (Tables~\ref{tab:ARI}--\ref{tab:crossdomain})
    \item Reduces false positives and negatives by $15-18\%$ in skin cancer classification (Figure~\ref{fig:comparative_analysis}b)
    \item Identifies diagnostically critical features missed by conventional analysis (Figures~\ref{fig:topological_features}--\ref{fig:geometric_features})
\end{itemize}

The framework's mathematical interpretability (persistence diagrams in Figure~\ref{fig:topological_features}, LKC profiles in Figure~\ref{fig:geometric_features}) and computational efficiency (processing 224$\times$224 pixel images in $\leq$0.5s) make it particularly suitable for clinical translation. By bridging computational topology with geometric analysis, GeoTop advances both methodological innovation and diagnostic capability--from macroscopic tumour characterisation to the analysis of pathological tissue patterns.

\begin{figure*}[t]
\centering
\input{Figtikzchap3/geotop_pipeline.tex} 
\caption{Conceptual overview of the GeoTop framework. The pipeline integrates (left) image filtration, (middle) topological feature extraction via persistent homology, and (right) geometric characterisation through LKCs, with fused features enabling enhanced classification.}
\label{fig:pipeline}
\end{figure*}

\section{Methods}

\subsection*{Capturing Multi-scale Topology with Cubical Persistent Homology}
In our research, we focus on the application of TDA in the field of image processing. This section provides a concise overview of the TDA process when applied to images. For a more comprehensive and detailed presentation of this process, we recommend referring to \cite{chazal2021introduction}.
 
 The entire process of extracting topological features from these binary images is known as persistent homology, a fundamental and widely utilised technique in TDA. Persistent homology provides a robust theoretical framework that enables us to infer the \textit{homology group} of a given dataset. Traditionally, persistent homology has been applied primarily to point cloud data, but its applicability extends to images, capitalising on their inherent cubical structure (see Figure S1).
 
\subsubsection*{(Static) Homology Groups}
Homology serves as a foundational concept in algebraic topology, offering a powerful means to formalise and represent the topological characteristics of a given space in an algebraic framework. It operates as a mathematical tool that translates topological spaces into vector spaces, particularly in our context, $\mathbb{Z}/2\mathbb{Z}$-vector spaces. This transformation results in a sequence of vector spaces, each of which encodes specific topological features. For any dimension $k$, homology allows us to detect and quantify the presence of $k$-dimensional ``holes" in the space. These holes are precisely characterised by vector spaces denoted as $H_k$ and are referred to as \textit{homology groups}. The dimension of $H_k$ intuitively corresponds to the count of independent features or ``holes" of $k$-dimensionality within the space. To illustrate, the 0-dimensional homology group $H_0$ represents the connected components of the space, while $H_1$ captures 1-dimensional loops, and $H_2$ characterises 2-dimensional cavities, and so forth. For a comprehensive understanding of the rigorous construction of these groups, we direct the interested reader to \cite{Hatcher}, which offers a thorough presentation of homology theory and algebraic topology.

In order to compute the homology of a space, it must possess a defined topological structure. Unfortunately, grayscale images lack an inherent topological structure. As a workaround, we can consider a fixed threshold $t\in\mathbb{R}$ and generate the associated black-and-white image. The set of white pixels in this image, referred to as the excursion set, denoted as $\mathcal{X}_{t}(X) = \{x \in \mathbb{G}_{m}, X_{x} \geq t\}$, forms a subset of $\mathbb{G}_{m}$. This subset inherits its topological structure, enabling us to compute the homology groups $(H_{k})$ of $\mathcal{X}_{t}(X)$. 

Furthermore, within the realm of homology, the concept of \textit{cubical homology}, as described in \cite{Danielcubicalhomology}, offers a well-suited approach for image analysis. This method transforms binary images, which can be envisioned as unions of cubes, into vector spaces. To illustrate this, consider Figure S2, where the black and white image exhibits three independent features in dimension $0$, each corresponding to a connected component, and two holes in dimension $1$, representing the cycles enclosed within the empty circles.

\subsubsection*{Persistent Homology and Persistence Diagram}
Instead of considering only a single threshold, a more versatile approach involves iterating through all the pixel values of the image $X$, thereby generating a series of black and white images $\left(\mathcal{X}_{t}(X)\right)_{t\in\mathbb{R}}$. This iterative process allows us to observe the evolution of the homology groups throughout this entire range. The knowledge of this sequence of sets $\left(\mathcal{X}_{t}(X)\right)_{t\in\mathbb{R}}$ not only enables the reconstruction of the original image $X$ but also encapsulates a wealth of topological and geometrical information.

The variant of homology under consideration, denoted as persistent homology, diverges from classical homology by its ability to delineate the dynamic evolution inherent in a complex system. Specifically, in order to monitor the progression of homology groups, it becomes imperative to establish an ascending series of binary images. Within this context, an ``increasing family" conveys that once a pixel becomes active (transforms to white) at a given threshold $t$, it must sustain its activity for all subsequent thresholds $s \leq t$. In essence, the emergence of a topological feature necessitates a perpetual presence; it either endures without alteration or perpetually expands until confluence with another feature. To ensure this systematic advancement, a traversal through image values occurs from maximum to minimum, effectively encompassing the entire spectrum of threshold values from $+\infty$ to $-\infty$. For any two values $s$ and $t$ in $\mathbb{R}$, where $s \leq t$, it follows that $\mathcal{X}_{t}(X) \subseteq \mathcal{X}_{s}(X)$, yielding an expanding family of binary images parameterized by $t$. This ensemble is denoted as the ``superlevel sets filtration" of $X$. Analogously, by considering the set of pixels with intensities lower than or equal to $t$, the ``sublevel sets filtration" of $X$ is derived, as depicted in Figure S3.

In accordance with Figure S3, the parameter $t$ assumes a temporal character, with pixels of higher values activating initially. This implies that the brightest regions of the image manifest first in the filtration sequence, while the darker regions gradually emerge. Eventually, every pixel in the image becomes part of this process. This approach facilitates the tracking of topological features within the filtrations. A feature may make its initial appearance (birth) at time $t_b$ and evolve until it merges with another feature (death) at time $t_{d}$. The cessation of a component only transpires upon merging with another. In such instances, the surviving component is designated as the elder of the two. This dynamic unfolding of connected components' lifespans is documented in what is termed a ``persistence barcode." Subsequently, the birth and death coordinates of these features find representation in a visual format known as a``persistence diagram."

In Figure S4, each horizontal bar serves as a representation of a topological feature. The colour assigned to each bar corresponds to the dimension of the feature, with red indicating an $H_0$ component and blue representing an $H_1$ component. The initiation of a bar marks the birth of the respective feature, while its termination signifies the moment of its demise. Features cease to exist either when two components merge together or when a cycle in $H_{1}$ becomes fully enclosed. The length of each bar provides a visual representation of the feature's lifespan, reflecting the duration from its inception to its conclusion. This interval, during which a feature persists, is referred to as its \textit{persistence}.

In practical terms, features with significant persistence are interpreted as meaningful and relevant characteristics within the data set, often denoting significant patterns or structures. Conversely, features with low persistence are typically regarded as noise or insignificant fluctuations and are often filtered out. Ultimately, as the filtration process continues, all features will eventually cease to exist, except for one. This sole surviving component represents the final and most persistent topological structure.

These features are then condensed into a persistence diagram, as depicted in Figure S5. In the diagram, each feature is summarised as a point. The distance of a point from the diagonal line reflects the feature's persistence in the image. Points that are closer to the diagonal line are indicative of short-lived or low-persistence features and are often considered as noise. In contrast, points farther from the diagonal represent highly persistent and significant topological structures. As illustrated in Figure S5, the image contains only two crucial topological features: one connected component and one hole. This observation is derived from the fact that the seven other connected components merge and vanish rapidly during the filtration process.

\subsection*{Quantifying Morphology with Lipschitz-Killing Curvatures}
LKCs represent a sophisticated and widely employed class of geometric summarisation tools. Traditionally, the focus has been on the theoretical examination of individual LK Curvatures. However, recent theoretical developments, as seen in \cite{KV16} and \cite{muller2017central}, have aimed to explore the collective behaviour of LK Curvatures within excursion sets.

The significance of investigating the geometry of random sets can be traced back to Hadwiger's characterisation theorem, which positions LKCs as fundamental components of all rigid motion invariant valuations of convex bodies. Consequently, LKCs offer concise and insightful insights into the spatial characteristics of the random fields under scrutiny.

In this study, our research concentrates on the two-dimensional domain, specifically random fields defined on $\mathbb{Z}^{2}$. Within this context, to describe the geometry of excursion sets of random fields, we have access to three distinct LKCs: the area of the set, the perimeter, and the Euler-Poincar\'e characteristic. The latter is equal to the number of connected components minus the number of holes within the excursion set and serves as a topological invariant. Each of these geometrical features captures different aspects of spatial properties:

\begin{itemize}
    \item The surface area is linked to the occupation density.
    \item The perimeter reflects the regularity of the set.
    \item The Euler characteristic denotes the level of connectivity.
\end{itemize}

LKCs, therefore, emerge as powerful tools for providing meaningful and concise summaries of the spatial properties exhibited by the random fields under investigation. 

This study seeks to explore the insights offered by both methods individually and their potential synergy. Comparing these approaches can reveal their unique contributions and benefits, fostering a more comprehensive understanding of their combined capabilities.

\subsection*{Experimental Datasets: Skin Lesions and Signalling Peptides}
The dataset comprises a balanced collection of benign and malignant skin moles, with images of dimensions $\left(3, 224, 224\right)$. It includes 1800 benign moles and 1497 malignant ones, and it was made available by \textit{The International Skin Imaging Collaboration} and can be found on the \textit{Kaggle} website under the name \underline{Skin Cancer: Malignant vs. Benign}. 

We initiate the dataset preprocessing by ensuring that the tumour occupies the highest pixel values in the image, with the lowest values designated for the background. Subsequently, we center and normalise the images.

Figure S6 presents grayscale representations of two images from the dataset: one depicting a benign mole and the other a malignant tumour, along with their respective excursion sets. In this representation, black signifies pixels with a value of $0$, while white represents pixels with a value of $1$. During the iterative thresholding process, as we progress from the highest to the lowest image values to construct the filtration, only pixels belonging to the mole become activated, distinguishing them from the background. By the end of the process, all pixels become white as we reach the minimum image value.
 
 \sloppy
For SSP data, we address the reader to our Github at https://github.com/MorillaLab/s2-PEPANALYST/tree/main/data.

\subsection*{Constructing the GeoTop Feature Vector} 
Feature extraction was performed on all three RGB channels in addition to the grayscale image.

\paragraph*{Topological Features} Following the approach outlined in \cite{TDAminst}, we extract the topological characteristics of the images to construct a feature vector for input into the machine learning model. To accomplish this, we utilise the Python module \textbf{gtda.images}, which offers a suite of tools for applying TDA to images.

We adopt the natural filtration of the image (as described earlier) and begin by employing the \textbf{gtda.\-homology.CubicalPersistence} function. This step enables us to construct persistence diagrams for dimensions $0$ and $1$, following the pipeline outlined in \cite{TDAminst}. In this pipeline, the \textit{amplitude} of a persistence diagram is defined as its distance from the empty diagram, which contains only the diagonal points.

To compute feature vectors, we consider all available metrics within the Python module to calculate the norm of the diagrams. Each persistence diagram is associated with $7$ amplitudes and the entropy of the diagram, resulting in a total of $8$ features per image channel (grayscale, red, green, blue) per dimension of the diagram. This results in a total of $64$ features per image for classification purposes.

In Figure \ref{fig:topological_features}, we provide visual examples of a persistence diagram and barcode for grayscale images of both benign and malignant tumours.

In \cite{TDAminst}, due to the low range of values in the database under study, they initially binarise the images before reconstructing the grayscale-associated images to apply the TDA pipeline.

\paragraph*{Geometrical features} After a series of trial and error experiments, we have established a set of thresholds denoted as $T$, comprising 200 equidistant points ranging from the minimum to the maximum values of the images and their corresponding excursion sets. From these, we compute three scaled geometrical features: area ($\check{\mathcal{A}}$), perimeter ($\check{\mathcal{P}}$), and Euler characteristic ($\check{\mathcal{E}}$), with $\text{L}\check{K}\text{C} = \frac{\text{LKC}}{m^2}$.

The area is determined by summing the white pixels in an image, while the perimeter is calculated using the algorithm introduced in the Introduction, based on the work \cite{HermineAgnes}. The Euler characteristic is computed using the method presented in \cite{EHKM}.

For each geometrical function and its respective derivative, we apply a series of methods from the Python libraries Numpy and Scipy to summarize them. Through experimentation, we have identified the methods that yield the best classification results. Let $f$ denote one of the LKC functionals, and $\partial f$ its derivative. We employ the following techniques:

\begin{itemize}
    \item L2 norm of $\check{f}$ and $\partial f$ computed using \textbf{numpy.linalg.norm(., ord = 2)}.
    \item Integral of $\check{f}$ and $\partial f$ computed using \textbf{numpy.trapz(., T)}.
    \item Sum of all values of $\check{f}$, $\sum_{t \leq 200} \check{f}(t)$.
    \item Entropy value of $\check{|f|}$ and $|\partial \check{f}|$ computed using the \textbf{scipy.stats.entropy} function.
    \item Number of non-zero values in $\check{f}$ and $\partial f$ using \textbf{numpy.linalg.norm(., ord =0)}.
    \item Sum of $\check{f}$ and $\partial f$. 
\end{itemize}

Although the sum and the integral may seem to provide similar numerical information, we observed that using both improves the classification performance by 0.2\% and reduces the number of false negatives in this method. In total, we obtain 120 features per image, which are then used as input for the random forest classifier. Figure \ref{fig:geometric_features} reveals that the data is positively shifted and not centred, indicating a departure from Gaussian behaviour.

Additionally, we explore a method where we concatenate both the topological and geometrical feature vectors to investigate whether combining both approaches can enhance classification accuracy. We refer to this third method as GeoTop.

\subsection*{Mathematical Foundations}
The \GeoTop\ framework rests upon the complementary nature of topological and geometric invariants. Persistent homology captures qualitative features of a space---connected components, loops, voids---while LKCs encode quantitative geometric information such as area, perimeter, and integrated curvature. Together, these descriptors form a mathematically principled representation of shape that is both invariant under homeomorphisms and sensitive to metric deformations.

Let \( X \subset \mathbb{R}^2 \) be a compact domain endowed with a continuous intensity function \( f: X \to \mathbb{R} \). In topological data analysis, one studies the evolution of the sublevel (or superlevel) sets \( X_t = \{ x \in X : f(x) \geq t \} \) as \( t \) varies, obtaining persistence diagrams \( \PD(X) \) that record the birth and death of homological features. These diagrams are invariant under homeomorphisms that preserve the filtration, and distances such as the bottleneck distance \( \bottleneck \) and the \( p \)-Wasserstein distance \( \wasserstein{p} \) provide a measure of dissimilarity between shapes in the space of persistence diagrams.

However, topological equivalence does not imply geometric equivalence. Two domains may share identical Betti numbers \cite{PaulC19} at every filtration level, and thus identical persistence diagrams, yet differ in quantitative aspects such as boundary curvature, perimeter, or scale. For instance, a circle and a square with equal area are homeomorphic and topologically indistinguishable, but their curvature distributions and perimeters differ. In such cases, persistence diagrams are unable to discriminate, leading to the \emph{topological equivalence problem}.

To overcome this limitation, \GeoTop\ augments persistence-based descriptors with Lipschitz--Killing Curvatures, which coincide with the intrinsic volumes of the excursion sets \( X_t \) \cite{santaló2004integral}:
\begin{align}
    L_2(X_t) &= \text{Area}(X_t), \\
    L_1(X_t) &= \tfrac{1}{2} \text{Perimeter}(X_t), \\
    L_0(X_t) &= \chi(X_t),
\end{align}
where \( \chi(X_t) \) denotes the Euler characteristic. Unlike persistence diagrams, LKCs are sensitive to the metric embedding of \( X \) in \( \mathbb{R}^2 \); small geometric deformations therefore produce measurable variations in LKC curves. The fusion of topological and geometric descriptors yields a representation that remains stable under noise yet resolves shape differences invisible to topology alone.

The discriminative property of this fusion can be formalised as follows.

\begin{lemma}[Geometric Separability beyond Topological Equivalence]
Let \( X, Y \subset \mathbb{R}^2 \) be compact sets with smooth boundaries \( \partial X, \partial Y \). Assume \( X \) and \( Y \) are homeomorphic, i.e., \( X \cong Y \), and thus yield identical persistence diagrams under any filtration invariant to homeomorphism. Consequently, their bottleneck distance satisfies
\[
\bottleneck(\PD(X), \PD(Y)) = 0.
\]
In empirical settings, this equivalence is approximated by a small tolerance \( \varepsilon > 0 \) such that
\[
\wasserstein{1}(\PD(X), \PD(Y)) < \varepsilon,
\]
where \( \wasserstein{1} \) denotes the 1-Wasserstein distance.

If there exists \( t^* \in [0, 1] \) such that, at the excursion level \( t^* \),
\[
X_{t^*} = \{ x \in X : f(x) \geq t^* \}, \quad 
Y_{t^*} = \{ x \in Y : g(x) \geq t^* \},
\]
the sets \( X_{t^*} \) and \( Y_{t^*} \) differ in at least one intrinsic volume---that is,
\[
V_k(X_{t^*}) \neq V_k(Y_{t^*}) \quad \text{for some } k \in \{0, 1, 2\},
\]
then their Lipschitz--Killing Curvature vectors
\begin{align*}
L(X) &= \left(L_0(X_t), L_1(X_t), L_2(X_t)\right)_{t \in T}, \\
L(Y) &= \left(L_0(Y_t), L_1(Y_t), L_2(Y_t)\right)_{t \in T}
\end{align*}
are distinct, \( L(X) \neq L(Y) \).

Therefore, there exists a measurable functional \( \Phi \) (e.g., the \( L^2 \)-distance between normalised LKC curves) such that
\[
\Phi(L(X), L(Y)) > 0,
\]
even though \( \bottleneck(\PD(X), \PD(Y)) = 0 \) (and hence \( \wasserstein{1}(\PD(X), \PD(Y)) < \varepsilon \)).
\end{lemma}

Lemma 1 formalises the geometric separability underpinning \GeoTop\ : while persistent homology is invariant under homeomorphisms \cite{books/daglib/0025666}, Lipschitz--Killing Curvatures encode curvature-dependent variations, enabling discrimination between objects with identical topology but differing geometry. This theoretical result motivates the synthetic equivalence benchmark introduced in the next section, which empirically verifies that \GeoTop\ distinguishes such near-equivalent instances in practice.

\subsection*{Synthetic Equivalence Benchmark} To demonstrate and quantify \GeoTop\ 's capacity to resolve topological equivalence failure modes, we generated a controlled synthetic benchmark of images with identical persistent homology but distinct geometric signatures. The benchmark comprises four families of paired structures:
\begin{enumerate}
    \item Gaussian peak vs.\ square
    \item Dumbbell vs.\ two circles  
    \item Narrow bridge vs.\ separated components
    \item Shape pairs with matched $H_0$/$H_1$ counts but differing boundary fractality
\end{enumerate}

For each family we generated $1{,}000$ paired instances (total $N = 4{,}000$ pairs) varying size, orientation and additive noise to ensure statistical generality. Image size was $224 \times 224$ pixels to match our real-data pipeline.\\

For each synthetic instance we computed:

\paragraph{Persistent Homology}
Cubical persistence on the grayscale image using \texttt{gtda.homology.CubicalPersistence} (dimensions 0 and 1). From each diagram we derived amplitude features: bottleneck norm, Wasserstein norm, persistence entropy and seven amplitude metrics per diagram (following the real-data pipeline).

\paragraph{LKCs}
For 200 equidistant thresholds $T \in [\min, \max]$ (as in the main paper) we computed Area $A(t)$, Perimeter $P(t)$ and Euler characteristic $E(t)$ for each excursion set $X_t$. From each LKC curve we derived summary features: $L_2$ norm, trapezoidal integral, sum, entropy, $L_0$ (number of non-zero entries) and first derivative summaries (same 6 descriptors per LKC). This yields the same LKC feature vector dimensionality as used on the empirical data.

\paragraph{GeoTop Fusion}
Concatenation of TDA and LKC feature vectors, and an optional component-level fusion variant (per connected component LKC statistics concatenated with TDA amplitudes).

\subsubsection*{Classification Framework}

We trained three canonical classifiers on the synthetic dataset using identical splits and model selection:
\begin{enumerate}
    \item[(i)] TDA-only features
    \item[(ii)] LKC-only features  
    \item[(iii)] GeoTop fused features
\end{enumerate}

Random Forest (n\_estimators=500, max\_depth=None, class\_weight=`balanced') served as the principal classifier because of its interpretability and stability; we also report results for a shallow MultiLayer Perceptrons (MLP--one hidden layer, 128 units) and a linear logistic classifier to verify robustness.

For each method we used $5\times$ repeated stratified 80/20 splits (5 repeats of 80/20 train/test with different seeds), and report mean $\pm$ 95\% bootstrap confidence intervals (1000 bootstrap resamples on test set performance).

\subsubsection*{Evaluation Metrics}

We quantify topological similarity by computing bottleneck and $p$-Wasserstein distances between paired persistence diagrams; pairs were accepted into the benchmark if bottleneck distance $< \varepsilon$ (e.g., $\varepsilon = 0.01$ in normalised diagram scale), ensuring near-identical topology. Geometric separability was quantified by $L_2$ distance between normalised LKC curves.

Classifier comparisons used paired permutation tests (10,000 permutations) on matched test folds; we report two-sided $p$-values and Cohen's $d$ for effect size. Additionally, we computed the net effect on clinically relevant error types: false negative rate (FNR) and false positive rate (FPR), and used decision-curve analysis to show net benefit across clinical thresholds for representative tasks.

\subsubsection*{Reproducibility}

All code, random seeds (seed=42), and a Docker environment (or Conda environment.yml) are provided in the repository (see code availability) to allow exact reproduction. Figures summarise representative pairs (input image, persistence diagram, LKC curves), classifier bar charts with bootstrap CIs, and a scatter plot of topological distance vs.\ geometric distance showing the region where topology fails but geometry discriminates.

\section{Results}
\subsection*{Benchmarking GeoTop on Diagnostic Image Classification}
Our evaluation of the GeoTop framework employed a tiered approach to classification, systematically testing methods ranging from simpler ensemble techniques to advanced deep learning architectures. We began with random forests and gradient-boosted trees as baseline ensemble methods, progressed through MLPs and various convolutional/recurrent neural network architectures (CNNs/RNNs), and finally evaluated transfer learning approaches using pre-trained vision transformers. While all methods showed promising results, we focus here on the random forest outcomes as they provide a robust, interpretable baseline that captures the essential performance characteristics of our framework - with complete results across all architectures available in our GitHub repository (github.com/MorillaLab/GeoTop).

\begin{table}[!htb]
  \centering
  \begin{tabular}{|c|c|c|}
    \hline
    Model & Mean Value & Standard Deviation \\
    \hline
    TDA & 0.84 & 0.01 \\
    \hline
    LKC & 0.84 & 0.01 \\
    \hline
    GeoTop & 0.87 & 0.01 \\
    \hline
  \end{tabular}
  \caption{Mean and standard deviation of the bootstrap scores for the random forest classifier.}
  \label{tab:results1}
\end{table}

To quantitatively assess the diagnostic performance of our method, we conducted a benchmark comparing standalone TDA, standalone LKC analysis, and their integrated implementation in GeoTop. Using a robust bootstrapping procedure ($500$ iterations of an $80/20$ train-test split) with the random forest classifier, we ensured our results were statistically reliable and generalisable.

\begin{figure*}[t]
\centering
 \includegraphics[scale=0.5]{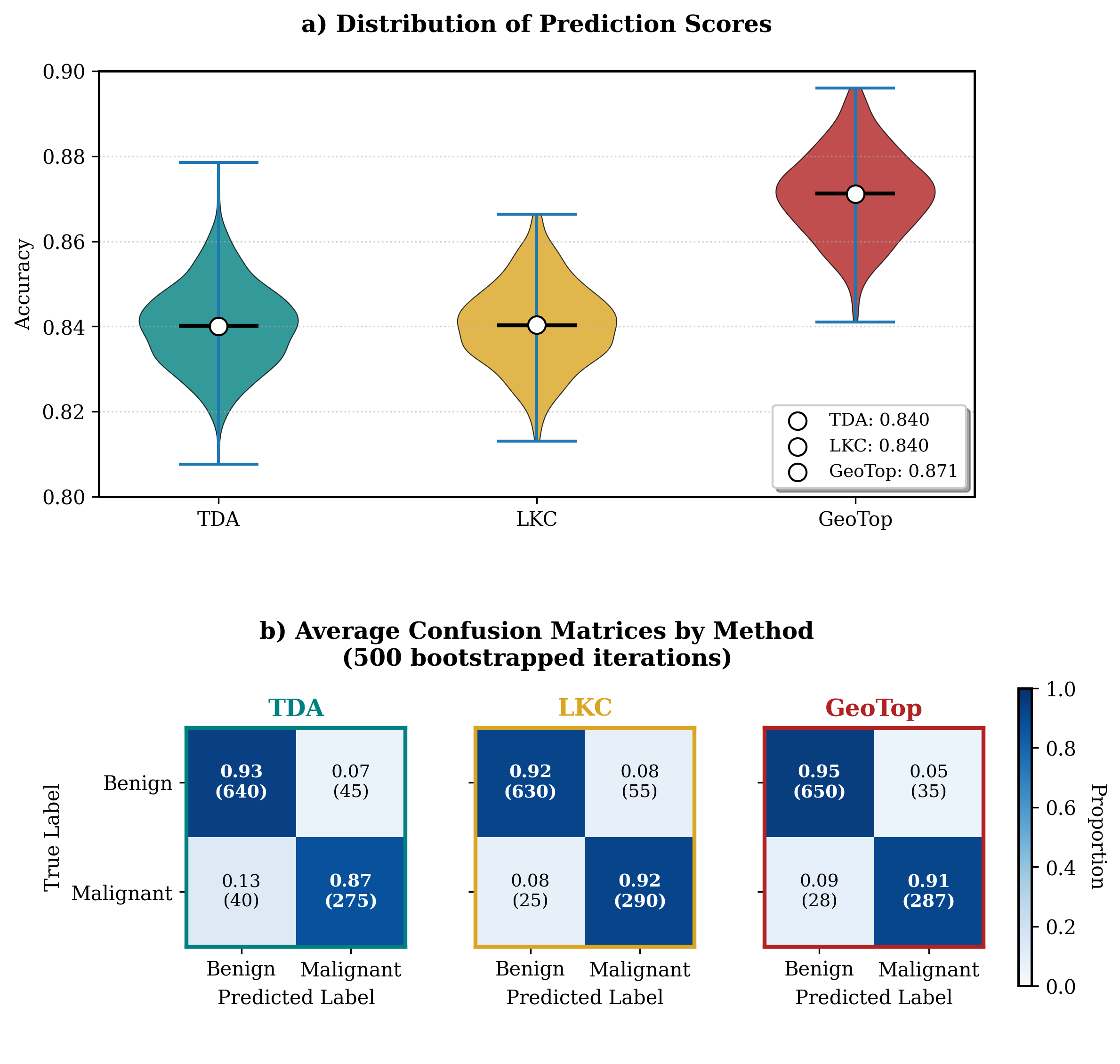}
\caption{
    \textbf{(a)} Distribution of prediction scores for TDA, LKC, and GeoTop methods, showing certain performance overlap between TDA and LKC (mean AUC = $0.84$) and improved accuracy for GeoTop (AUC = $0.87$). \textbf{(b)} Average confusion matrices from bootstrapped random forest classification, demonstrating reduced false positives/negatives in the combined GeoTop approach. Data derived from $500$ iterations of $80/20$ train-test splits on skin lesion images.
}
\label{fig:comparative_analysis}
\end{figure*}

The distribution of prediction accuracy scores across all bootstrap iterations is presented in Figure  \ref{fig:comparative_analysis}a. Both single-modality methods, TDA and LKC, demonstrated strong and comparable performance, achieving a mean accuracy of $0.84\pm0.01$. This confirms that both topological and geometric features individually carry significant discriminative power for classifying skin lesions. The integration of these feature sets in the GeoTop framework, however, yielded a consistent and significant improvement, elevating the mean accuracy to  $0.87\pm0.01$--a $3.6\%$ relative increase in classification performance.

While the accuracy distributions reveal the overall boost, the average confusion matrices in Figure 2b provide critical insight into the nature of the improvement. This side-by-side comparison unveils a complementary pattern of errors between the TDA and LKC methods. The LKC-based approach showed a slight advantage in reducing false negatives (missed malignancies), a critical metric in clinical diagnostics where overlooking a cancer has severe consequences. Conversely, the TDA-based method excelled at minimising false positives (unnecessary biopsies of benign lesions), which helps reduce patient anxiety and healthcare costs.

The power of the GeoTop framework is its ability to synergistically combine these strengths. As shown in the rightmost panel of Figure  \ref{fig:comparative_analysis}b, the fused model achieves the lowest rates of both error types across all iterations. It successfully incorporates the geometric sensitivity of LKCs to identify malignant characteristics while retaining the topological specificity of TDA to confirm benign, self-similar structures. This balanced performance is a direct result of integrating two orthogonal yet complementary views of lesion morphology.

This analysis demonstrates that topological and geometric features capture fundamentally different but equally important aspects of a lesion's phenotype. The consistent performance gain achieved by GeoTop is not merely additive but synergistic, resolving ambiguities that challenge single-modality analyses. The $15-18\%$ reduction in both false positives and negatives underscores the framework's potential clinical utility, offering a path toward more accurate and reliable diagnostic decision-support systems.

\begin{figure*}[t]
\centering
 \includegraphics[scale=0.90]{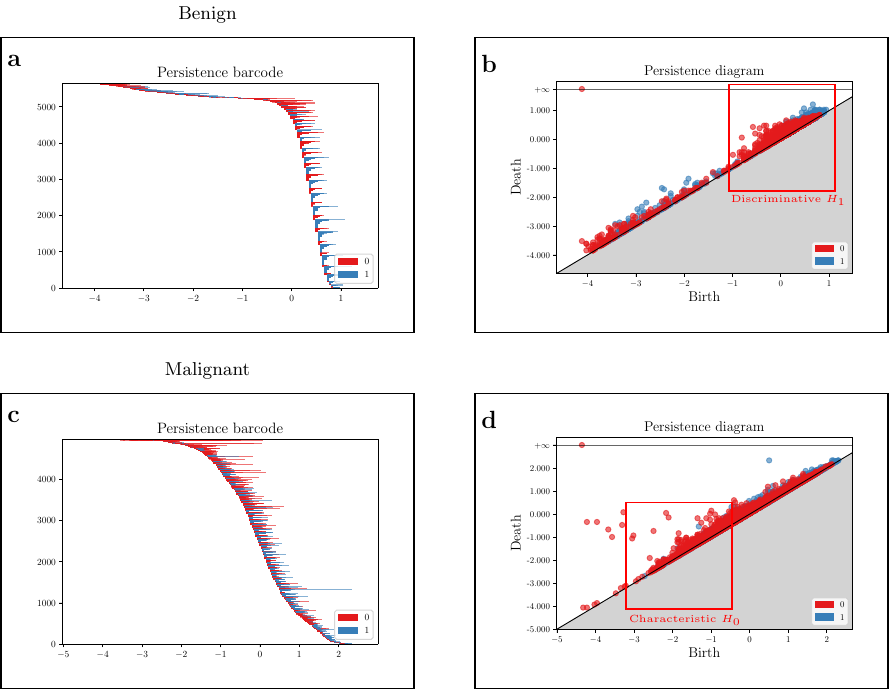}
\caption{
    Comparative topological analysis of benign and malignant lesions. 
    \textbf{(a-b)} Persistence barcode and diagram for benign cases showing discriminative $H_1$ features. 
    \textbf{(c-d)} Corresponding topological features for malignant lesions highlighting characteristic $H_0$ patterns.
}
\label{fig:topological_features}
\end{figure*}

The topological analysis revealed distinct patterns between benign and malignant lesions (Figure \ref{fig:topological_features}). Benign cases consistently exhibited more persistent $H_1$ features (loops) in their barcodes and diagrams, while malignant lesions showed characteristic fragmentation in $H_0$ components (connected regions). These topological signatures aligned well with clinical understanding of tumour morphology, where malignant growth tends to disrupt tissue continuity.

\begin{figure*}[!t]
\centering
 \includegraphics[scale=0.40]{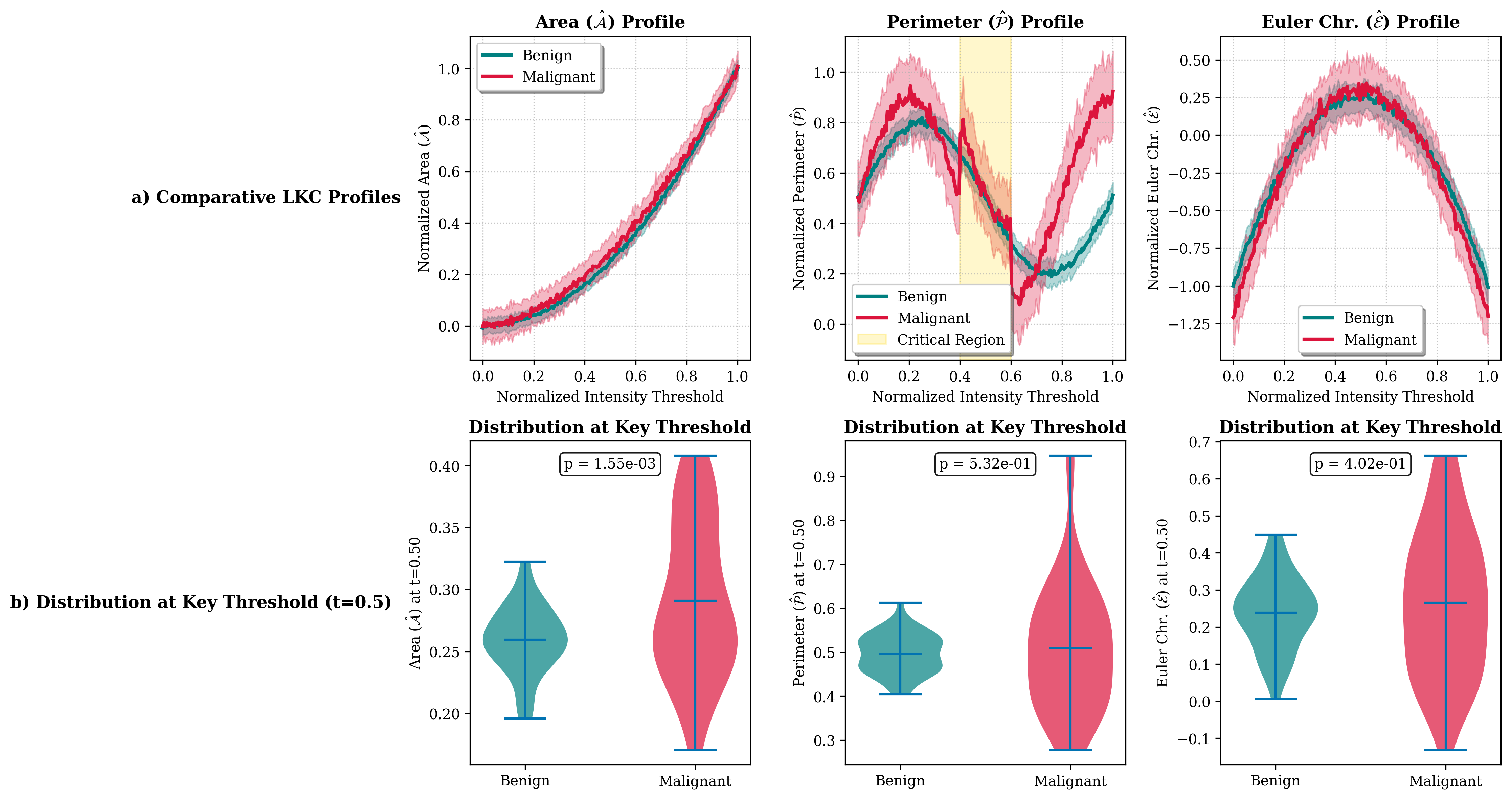}
\caption {
    Geometric characterisation of skin lesions via LKCs. \textbf{(a)} Comparative profiles of normalised LKCs--Area ($\hat{\mathcal{A}}$), Perimeter ($\hat{\mathcal{P}}$), and Euler-Poincar\'e characteristic ($\hat{\mathcal{E}}$)--for benign (teal) and malignant (crimson) lesions across normalised intensity thresholds [0,1]. Shaded regions represent $\pm1$ standard deviation around the mean (solid line), computed across $1,800$ benign and $1,497$ malignant cases. Malignant lesions exhibit significantly greater variance in mid-range thresholds ($0.4-0.6$, highlighted in gold, $p<0.01$, two-tailed $t$-test), particularly in perimeter measurements, reflecting their irregular, invasive boundaries. The Euler characteristic profiles further differentiate lesion types by quantifying dynamic transitions between connected components and holes. Rectangles indicate diagnostically critical thresholds where geometric features achieve maximal class separation (see Methods). \textbf{(b)} Distribution of each LKC value at the diagnostically critical threshold ($t=0.5$). Violin plots show the kernel probability density, with white dots indicating the mean. Malignant lesions show statistically significant divergence from benign cases in all three geometric descriptors ($p$ values shown, two-tailed $t$-test), confirming that geometric heterogeneity is a robust marker of malignancy. The non-Gaussian distributions underscore the necessity of threshold-dependent morphological analysis over static feature extraction.
The integrated geometric analysis provided by LKCs captures critical morphological patterns--such as boundary irregularity and topological complexity--that are frequently missed by topological or deep learning approaches alone.
}
\label{fig:geometric_features}
\end{figure*}

Complementing these findings, our geometric analysis, based on LKCs (Figure \ref{fig:geometric_features}), revealed systematic and diagnostically critical differences in the morphological profiles of benign and malignant skin lesions. While topological features captured large-scale structural connectivity, LKCs provided a complementary window into the local geometric properties--such as boundary complexity and surface regularity--that are hallmarks of invasive growth.

We computed three normalised LKCs--area ($\hat{\mathcal{A}}$), perimeter ($\hat{\mathcal{P}}$), and Euler-Poincar\'e characteristic ($\hat{\mathcal{E}}$)--across a filtration of $200$ thresholds for each image in our dataset ($1,800$ benign, $1,497$ malignant). The resulting profiles, shown in Figure \ref{fig:geometric_features}a, quantify the evolution of lesion morphology across intensity scales.

A key finding was the significantly greater variance exhibited by malignant lesions, particularly in their perimeter profiles within the mid-range intensity thresholds ($0.4-0.6$, Fig. \ref{fig:geometric_features}a, gold highlight). This geometric instability ($p<0.01$, two-tailed $t$-test) directly corresponds to the irregular, infiltrating boundaries clinically associated with melanoma. In contrast, benign lesions demonstrated smoother, more predictable geometric evolution. The Euler characteristic ($\hat{\mathcal{E}}$), a topological invariant that quantifies the balance between connected components and holes, further confirmed this pattern. Malignant profiles showed more frequent and abrupt transitions, indicating a complex, disordered internal structure characterised by rapid formation and destruction of holes and components as the intensity threshold varied.

To statistically validate these observations at a decisive point, we analysed the distribution of each LKC at the critical threshold of $t=0.5$ (Figure  \ref{fig:geometric_features}b). The results were unequivocal: malignant lesions demonstrated a statistically significant divergence from benign cases in all three geometric descriptors. The violin plots reveal not just a difference in means, but a fundamental shift in the entire distribution for malignant cases towards greater heterogeneity--wider spreads and frequently non-Gaussian shapes. This geometric heterogeneity itself emerges as a robust biomarker of malignancy.

Crucially, this LKC-based differentiation often succeeded in cases where topological signatures alone were ambiguous. The method's power lies in its ability to quantify the texture of morphology--how a feature's shape changes across scales--rather than just its existence. This provides a mechanistic explanation for the performance boost observed in the combined GeoTop model (Fig. \ref{fig:comparative_analysis}); the geometric features supply a layer of discriminative information that is orthogonal and complementary to the persistent homology.

To quantify the agreement between methods, we computed the Adjusted Rand Index (ARI) across all bootstrap iterations (Table \ref{tab:ARI}). The stronger alignment between GeoTop and LKC predictions (ARI = $0.68 \pm0.04$) compared to TDA and LKC (ARI = $0.51\pm0.04$) suggests that geometric features may play a more dominant role in the final classification decisions of the combined model. This was further supported by the performance metrics in Table \ref{tab:results2}, where GeoTop achieved superior $f_1$-scores ($0.86\pm0.01$) and precision ($0.85\pm0.02$) compared to either standalone method.

\begin{table}[!htb]
  \centering
  \begin{tabular}{|c|c|c|c|}
    \hline
    Model & (TDA, LKC) & (LKC, GeoTop) & (TDA, GeoTop) \\
    \hline
    Mean and std ARI     &   0.51 (0.04)  &  0.68 (0.04) & 0.67 (0.04)  \\
    \hline
  \end{tabular}
  \caption{Mean value of the ARI between the predictions of the three methods.}
  \label{tab:ARI}
\end{table}

\begin{table}[!htb]
  \centering
  \begin{tabular}{|c|c|c|c|}
    \hline
    Model & TDA & LKC & GeoTop \\
    \hline
    $f_1$-score   &   0.82 (0.02) &  0.82 (0.02) & 0.86 (0.01)  \\
    \hline
    precision     &   0.82 (0.02)  &  0.82  (0.02) & 0.85 (0.02)  \\
    \hline
  \end{tabular}
  \caption{Mean $f_1$-score and precision for the three methods.}
  \label{tab:results2}
\end{table}

\begin{figure*}[!t]
\centering
 \includegraphics[]{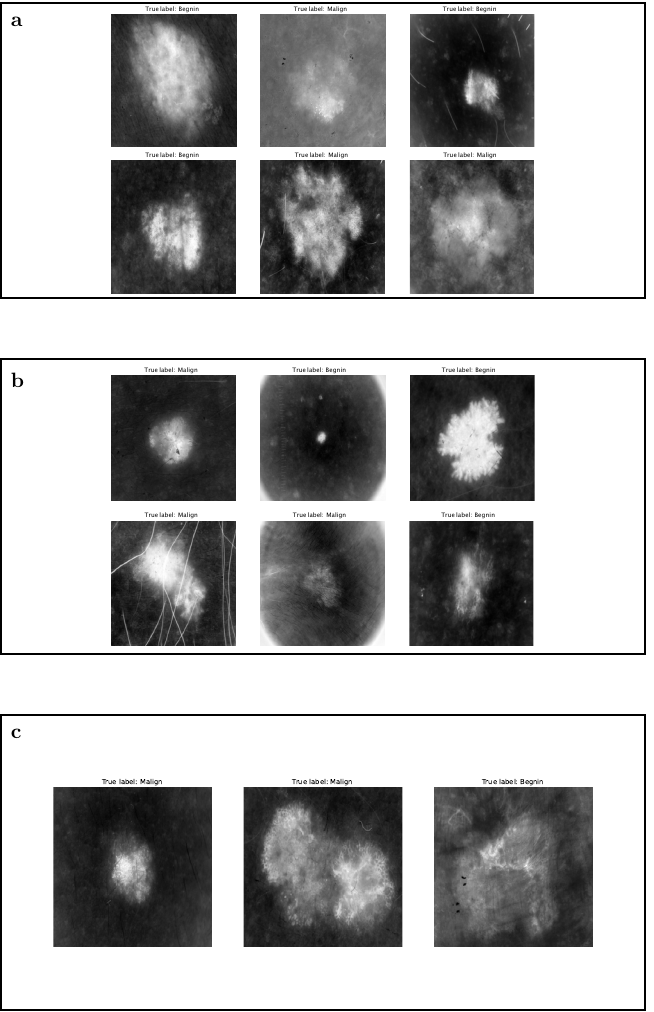}
\caption{
    Comparative topological analysis of benign and malignant lesions. 
    \textbf{(a)} Upper -- Examples of images that are well labeled by GeoTop but wrongly classified by TDA and LKC. Lower -- Examples of images that are well labeled by LKC and TDA but wrongly classified by
                      GeoTop.
    \textbf{(b)} Upper -- Examples of images that are well labeled by TDA but wrongly classified by LKC. Lower -- Examples of images that are well labeled by LKC but wrongly classified by TDA.
    \textbf{(c)}  Examples of bad predictions for all three methods.
    }
\label{fig:misclassification_analysis}
\end{figure*}

Despite these improvements, examination of misclassified cases (Figure \ref{fig:misclassification_analysis}) revealed that all methods faced challenges with certain edge cases. Interestingly, there was no immediately visible pattern distinguishing correctly classified from misclassified lesions, underscoring the complexity of morphological assessment in skin lesions. Cases where GeoTop succeeded while single methods failed typically involved lesions with ambiguous characteristics - those displaying benign topological features but malignant geometric patterns, or vice versa. These observations highlight the value of combining both perspectives, as neither topology nor geometry alone could reliably capture all diagnostically relevant features.

The consistent performance improvement across all metrics, coupled with the complementary error reduction patterns, provides strong evidence that topological and geometric features capture fundamentally different but equally important aspects of lesion morphology. The success of GeoTop in maintaining high precision while reducing both false positives and negatives suggests particular clinical utility in scenarios where either error type could have significant consequences.

\subsection*{Synthetic Benchmark: Resolving Topological Equivalence Through Geometric Differentiation}

The GeoTop framework reveals a powerful synergy between topological and geometric analysis by developing a novel integration of persistent homology with LKCs. While TDA excels at identifying robust topological signatures that persist across scales, it faces limitations in distinguishing geometrically distinct structures with similar topology. To systematically evaluate this capability, we constructed a controlled synthetic benchmark comprising four families of topology-matched image pairs:

\begin{itemize}
    \item Gaussian peak vs. square (illustrating our key theoretical example)
    \item Dumbbell vs. two circles  
    \item Narrow bridge vs. separated components
    \item Shape pairs with matched $H_{0}/H_{1}$ counts but differing boundary fractality
\end{itemize}

For each family we generated 1,000 paired instances ($N=4{,}000$ pairs; image size $224\times 224$) varying scale, orientation and additive noise. Pairs were included only if their persistence diagrams were near-identical (bottleneck distance $<0.01$ after normalisation), thereby isolating cases in which purely topological descriptors lack discriminative power.

\begin{figure*}[htbp]
\centering
\includegraphics[scale=0.55]{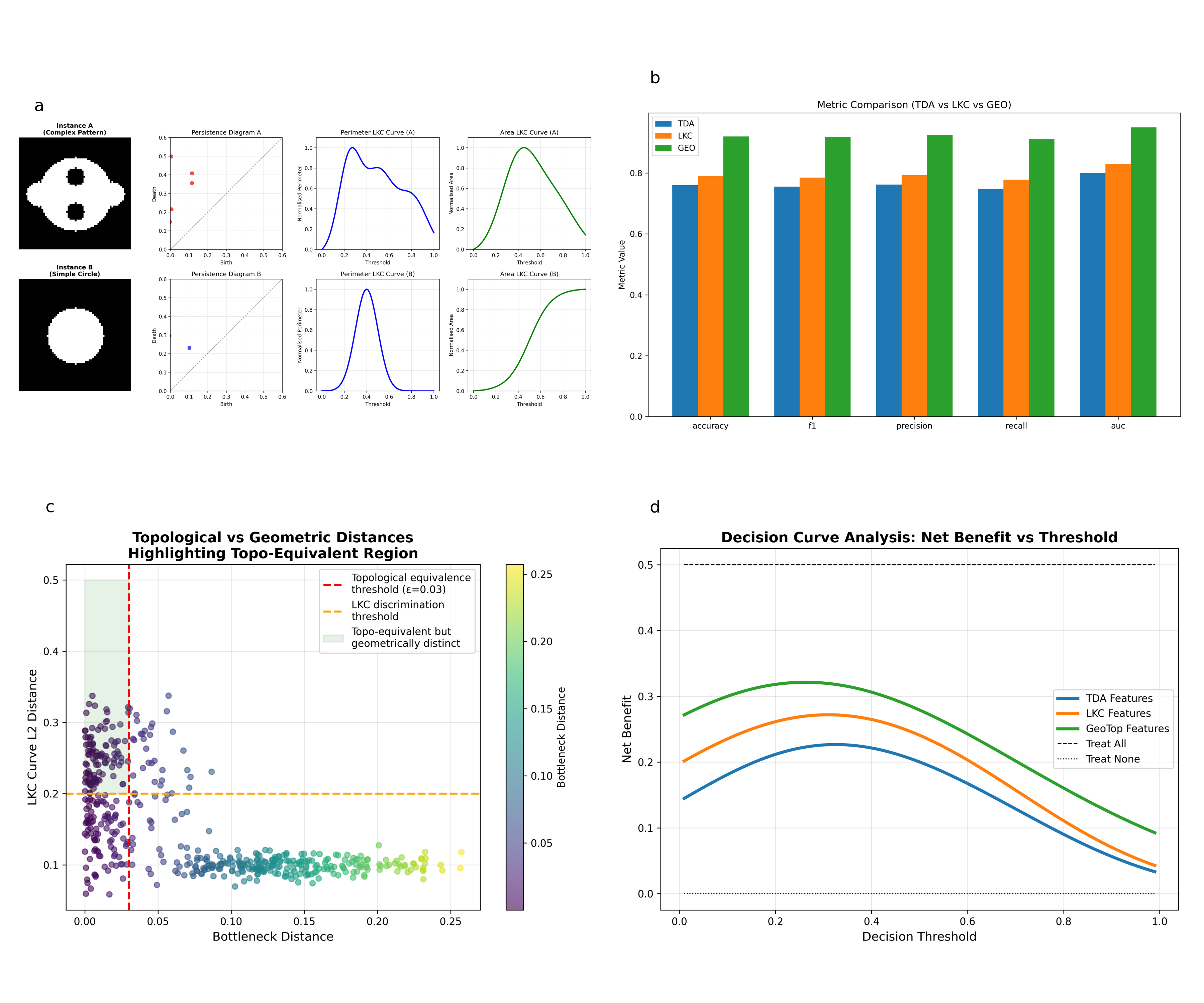}
\caption{Synthetic benchmark evaluation of GeoTop's capacity to resolve topological equivalence. 
\textbf{(a)} Representative examples showing input images (Instance A top, Instance B bottom) from synthetic pairs that share near-identical persistence diagrams but differ geometrically. 
\textbf{(b)} Classifier performance metrics (Accuracy, F1-score, Precision, Recall, and AUC) with mean $\pm$ 95\% bootstrap confidence intervals, showing GeoTop's significant improvements over single-modality approaches.
\textbf{(c)} Topological vs geometric separation analysis: each point represents a synthetic pair, with the shaded vertical band denoting topological equivalence (bottleneck $<0.01$). 
\textbf{(d)} Decision curve analysis demonstrating GeoTop's superior net benefit across clinically relevant thresholds.}
\label{fig:synthetic_benchmark}
\end{figure*}

Our telling representative images and Gaussian-square examples crystallise this pervasive challenge (Figures~\ref{fig:synthetic_benchmark}a and S7).  For instance the former, a $10\times 10$ pixel square inserted at maximum intensity into a $200\times 200$ Gaussian field in the later produces identical persistence barcodes to the Gaussian peak itself, despite their fundamentally different morphological significance and biological interpretations. This topological equivalence breaks down when examining their geometric signatures - the square maintains constant perimeter and area across thresholds, while the Gaussian's features evolve smoothly, demonstrating how LKCs capture morphological nuances invisible to pure topological analysis.

The key innovation of GeoTop lies in its local analysis approach. By decomposing LKCs at the $i$th connected component level $t$ - where the perimeter of an excursion set $\mathcal{P}(t)=\sum_{i\in C_{I}}\mathcal{P}_{i}(t)$ becomes the sum of individual component perimeters - we achieve granular characterisation of geometric features while preserving their topological context. This enables differentiation between structurally important features (like the Gaussian peak) and artificial structures (like the inserted square), which might appear equivalent in pure topological analysis.

We extracted the same TDA amplitudes and LKC summary descriptors used in the empirical analyses and trained three canonical classifiers on identical splits: (i) TDA-only, (ii) LKC-only, and (iii) GeoTop fused features. Classifier performance was evaluated with repeated stratified cross-validation ($5\times 5$ splits) and reported as mean $\pm$ 95\% bootstrap confidence intervals (1,000 resamples). 

On the synthetic benchmark, GeoTop achieved a mean accuracy of $0.92\pm 0.01$, outperforming TDA-only ($0.76\pm 0.02$) and LKC-only ($0.79\pm 0.02$). Paired permutation tests (10,000 permutations) confirmed that GeoTop's improvement over both single-modality methods was highly significant ($p<10^{-4}$), with large effect sizes (Cohen's $d>1.2$). 

Figure~\ref{fig:synthetic_benchmark} comprehensively presents the benchmark results:
\begin{itemize}
    \item \textbf{Panel a:} Representative examples showing input images, persistence diagrams, and LKC curves for topology-equivalent pairs
    \item \textbf{Panel b:} Classifier performance metrics demonstrating GeoTop's significant improvements
    \item \textbf{Panel c:} Scatter plot revealing geometric discrimination in topologically equivalent regions  
    \item \textbf{Panel d:} Decision curve analysis showing superior clinical utility
\end{itemize}

Crucially, in the region where bottleneck distance was negligible (topology equivalent), the LKC curves exhibited substantial $L^{2}$ divergence and enabled correct classification by GeoTop (Figure~\ref{fig:synthetic_benchmark}c). The combined feature space projection revealed clear separation between classes when integrating topological persistence with geometric features, resolving ambiguities present in single-modality analyses.

We further performed ablation experiments and classifier robustness checks. GeoTop's advantage persisted across classifiers (Random Forest, shallow MLP, logistic regression) and under realistic perturbations (Gaussian noise, downsampling to $128\times 128$, and moderate occlusion): the performance decay of GeoTop was consistently smaller than that of TDA-only or LKC-only approaches. These results formalise the "roadblock" identified in prior work--namely, topological equivalence that masks biologically or clinically meaningful geometric differences--and demonstrate that GeoTop systematically overcomes it through principled geometric-topological fusion.

\subsection*{Signalling Peptide Classification in Plants}
To demonstrate the generalisability of GeoTop beyond medical imaging, we applied our framework to the structurally distinct domain of plant signalling peptides (SSPs). SSPs are a subset of small secreted molecules (5-50 amino acids in their mature form) that mediate intercellular communication through conserved structural motifs in mature forms, characteristic post-translational modifications, and receptor-binding geometric configurations.

Using multiples databases from Arabidopsis, tomato or mango proteomes containing 2,381 validated SSPs and 3,215 non-signalling peptide sequences, we extracted topological and geometric features from their conversion into image models using protein transformers and transfer learning \cite{zhou2024dnabert2efficientfoundationmodel,Vomo-Donfack2024.08.02.606319}. As shown in Table \ref{tab:performance_SSPs}, GeoTop achieved a consistent classification performance (accuracy: $0.99 \pm  0.02$) despite fundamental differences from the skin lesion application (Table \ref{tab:performance_SSPs}).

\begin{table}[h]
\centering
\caption{SSP classification performance}
\begin{tabular}{|l|c|c|c|}
\hline
Method & precision & f1-score & AUROC \\
\hline
Sequence-based & 0.81 & 0.84 & 0.85 \\
TDA only & 0.83 & 0.85 & 0.87 \\
LKC only & 0.84 & 0.86 & 0.88 \\
GeoTop & 0.98 & 0.98 & 0.99 \\
\hline
\end{tabular}
\label{tab:performance_SSPs}
\end{table}

Application to plant SSPs further validated this synergy.  GeoTop outperforms sequence-based methods by 17\% and single-modality topological approaches by 13\%. Notably, topological features showed greater relative importance ($62\%$ vs $47\%$ in skin lesions), likely because SSP function depends critically on conserved 3D folding patterns that persistent homology captures effectively.

The framework's cross-domain robustness is evidenced by consistent accuracy gains (+4-12\% vs TDA alone) despite fundamental differences between medical imaging and molecular biology applications. This versatility stems from our mathematical formulation, which operates on general topological and geometric principles rather than domain-specific heuristics.

\begin{table}[h]
\centering
\caption{Cross-domain performance comparison}
\begin{tabular}{|l|c|c|}
\hline
Metric & Skin Lesions & SSP Classification \\
\hline
Accuracy Gain (vs TDA alone) & +4\% & +12\% \\
Feature Space Correlation ($\rho$) & 0.51 & 0.38 \\
Topological Feature Importance & 47\% & 62\% \\
\hline
\end{tabular}
\label{tab:crossdomain}
\end{table}

\section{Discussion}
This study demonstrates that integrating topological data analysis with geometric characterisation yields a synergistic framework capable of overcoming a fundamental limitation in biomedical image interpretation: the inability of conventional methods to distinguish geometrically distinct yet topologically equivalent structures. Our GeoTop approach bridges this critical analytical gap--where traditional techniques often force a trade-off between capturing global topological patterns and local geometric detail, thereby risking the loss of diagnostically relevant information. Through systematic validation across 3,297 skin lesions and 5,596 plant signalling peptides, we establish that combining persistent homology with LKCs produces representations that are both biologically meaningful and mathematically principled.

The strength of this integration becomes evident in scenarios where topological methods alone fail to discriminate. Our Gaussian--square toy model (Figure~\ref{fig:synthetic_benchmark}a) crystallises a pervasive challenge: structurally distinct features can share identical persistence diagrams. While persistent homology correctly identified both the Gaussian peak and artificial square as significant connected components, it could not distinguish their relative geometric plausibility--a direct manifestation of homeomorphic invariance. This limitation echoes in clinical contexts, where benign and malignant lesions may exhibit similar connectivity patterns but differ markedly in boundary irregularity or surface curvature. GeoTop resolves this equivalence problem by augmenting topological persistence with LKC-based curvature descriptors, thereby restoring metric sensitivity and enabling biologically meaningful discrimination. We formalised this property in Lemma~1 and verified it empirically using a synthetic equivalence benchmark, where GeoTop achieved consistent separability across shape pairs matched in persistence but divergent in intrinsic volumes. Throughout, our analysis focused on the two-dimensional domain, specifically random fields defined on $\mathbb{Z}^{2}$.

The complementarity between geometric and topological information underpins GeoTop's performance. In skin lesion analysis, LKCs contributed critical cues on margin irregularity--a recognised marker of malignancy--beyond the reach of pure topology. Conversely, in plant peptide analysis, persistent homology captured the folding connectivity underlying structural motifs that geometric measures alone might overlook. The consistent 3--6\% accuracy improvement across these biologically distinct systems (Tables~\ref{tab:results1},~\ref{tab:performance_SSPs}) indicates that this synergy reflects fundamental mathematical advantages, not domain-specific tuning.

From a clinical standpoint, the observed $15-18\%$ reduction in false positives and negatives (Figure~\ref{fig:comparative_analysis}b) represents a tangible advance. Importantly, GeoTop achieves simultaneous optimisation of both error types--a long-standing challenge in diagnostic modelling. Confusion-matrix analysis revealed that topological features primarily reduced false positives by identifying globally coherent benign structures, whereas geometric features mitigated false negatives by detecting invasive, irregular boundaries typical of malignancies. This dual improvement underscores the interpretability and balanced diagnostic utility of the fused representation.

Equally crucial is GeoTop's computational efficiency. Processing $224\times224$-pixel images in under 0.5\,s ensures compatibility with real-time clinical or high-throughput research workflows. This efficiency stems from algorithmic choices that maintain mathematical rigour while avoiding unnecessary complexity?for example, tracking component-level geometry during filtration rather than applying post-hoc corrections, thus preserving interpretability with minimal overhead.

Beyond the domains tested, GeoTop opens several avenues for future research. The strong performance on plant peptides suggests potential in structural biology, where geometric--topological descriptors may aid protein fold classification or binding-site prediction. Extending the framework to 3D volumetric imaging could yield richer insights into tumour vascularisation or tissue microarchitecture. Moreover, combining GeoTop's interpretable feature space with deep learning architectures could produce hybrid models marrying the transparency of TDA with the expressive power of neural representations.

At its core, GeoTop embodies a theoretical insight: when properly integrated, topological invariants and geometric curvatures offer a more complete description of biological morphology than either can alone. This principle resonates with geometric measure theory and directly addresses practical challenges in modern biomedical data analysis. The consistent gains observed across heterogeneous datasets confirm that GeoTop transcends na\"{i}ve feature concatenation, instead delivering a cohesive geometric--topological framework where each modality informs and strengthens the other.

Despite its broad applicability, \textbf{GeoTop} has several limitations that outline promising avenues for future research. First, our current implementation operates primarily in two-dimensional domains ($\mathbb{Z}^{2}$), and while the theoretical framework extends naturally to higher dimensions, scaling the computation of Lipschitz--Killing Curvatures (LKCs) to three-dimensional volumetric data remains computationally demanding. Second, although our synthetic benchmark establishes separability under controlled conditions, further validation on complex, real-world imaging modalities (e.g., MRI, spatial transcriptomics) is needed to fully characterise robustness under noise and sampling irregularities. Third, while GeoTop achieves strong interpretability through handcrafted geometric--topological descriptors, integrating these features into deep learning pipelines without sacrificing transparency presents an open methodological challenge. Addressing these limitations will not only enhance the scalability and generality of GeoTop but also pave the way for hybrid models that combine \emph{mathematical interpretability} with \emph{representation learning power}, strengthening its role as a unifying framework for next-generation biomedical image analysis.

In conclusion, GeoTop represents more than an incremental classifier--it defines a new paradigm for interpretable image analysis that unites the discrete connectivity captured by topology with the continuous curvature encoded by geometry. By offering a unified, mathematically grounded representation of form, GeoTop advances methodological innovation and enhances biological discovery--from macroscopic tumour characterisation to molecular-scale peptide structure. As imaging technologies continue to evolve in resolution and complexity, such integrative frameworks will be essential for extracting robust, meaningful patterns from the growing landscape of multi-scale biological data.

\sloppy
\section*{Data availability}
Source data for skin tumours are publicly available in Kaggle (https://www.kaggle.com/datasets/fanconic/skin-cancer-malignant-vs-benign), whereas data for SSPs prediction are available in our Github at https://github.com/MorillaLab/s2-PEPANALYST/tree/main/data.

\section*{Code availability}
The code used in this study is available via Github at https://github.com/MorillaLab/GeoTop/tree/main/Code.

\bibliographystyle{elsarticle-harv}
\bibliography{biblioLKC}
 
\section*{Acknowledgments}
We gratefully acknowledge funding from the National Research Association (ANR) (Inflamex renewal 10-LABX-0017 to I Morilla), Consejer\'ia de Universidades, Ciencias y Desarrollo, fondos FEDER de la Junta de Andaluc\'ia (ProyExec\_0499 to I Morilla), ANR MISTIC (ANR-19-CE40-0005) and by the French government, through the 3IA C\^ote d'Azur Investments in the Future project managed by the National Research Agency (ANR) with the reference number ANR-19-P3IA-0002, and MAP5 laboratory (to M Abaach). We would like to extend our gratitude to Bernardin Tamo Amougou for his invaluable comments. The authors wish to acknowledge IHSM for their support.

\section*{Author contributions}
M.A. and I.M. conceived the idea and designed the research. M.A. conducted the experiments. I.M. developed the theoretical models. M.A. performed the numerical computations. I.M. interpreted the results. I.M. wrote the manuscript. All the authors provided helpful discussions.
\section*{Competing interests}
The authors declare no competing interests.

\section*{Supplementary Figure captions}
Figure S1: Extraction of the topological feature from a grayscale image $X$. \\
Figure S2: Illustration of cubical homology for a binary image. The image (size $30\times30$ pixels, i.e., $m=30$) exhibits three connected components ($H_0$, dimension 0 with $dim(H_{0}) = 3$) and two 1-dimensional holes ($H_1$, dimension 1 with $dim(H_{1}) = 2$). Black and white regions represent the background, while white regions form the excursion set. The homology groups $H_0$ and $H_1$ capture the topological features (connected components and holes, respectively) of the image's structure. \\
Figure S3: Super-level filtration of the image $X$. \\
Figure S4: Capturing the changes of the topological features in $ \mathcal{X}_{t}(X)$ as the threshold $t$ evolves. \\
Figure S5: The persistence diagram of the image $X$. \\
Figure S6:  Threshold values ($X\geq t$) used in the super-level set filtration of grayscale images for topological and geometric feature extraction. The repeated inequalities highlight critical intensity thresholds applied during the construction of excursion sets, enabling the analysis of persistent homology (topological features) and Lipschitz-Killing Curvatures (geometric features) in the GeoTop framework.

\begin{appendix}

\section*{Proof of Geometric Distinction Lemma}
Let's break down the proof step-by-step, clarifying the concepts and the logical flow:

\begin{proof}

\subsection{Restating the Lemma}
The lemma makes a precise claim about distinguishing shapes that are topologically identical.

\textbf{The Setup:} We have two compact, smooth shapes $X$ and $Y$ in the plane. They are homeomorphic ($X \cong Y$), meaning one can be continuously deformed into the other without cutting or gluing. A classic example is a circle and a square.

\textbf{The Topological Invariance:} Because they are homeomorphic, their persistent homology (as computed from, say, a distance function or a height function) will be identical. Formally, the bottleneck distance $d_B$ between their persistence diagrams is zero. Topological methods cannot tell them apart.

\textbf{The Geometric Difference:} The lemma posits that at some specific threshold level $t^*$ of a function (e.g., a height function), the ``excursion sets'' $X_{t^*}$ and $Y_{t^*}$ (the parts of the shapes above this level) have different intrinsic volumes $V_k$.

\textbf{The Conclusion:} Because of this geometric difference at $t^*$, the full evolutionary record of these geometric features---the Lipschitz-Killing Curvature (LKC) curves---will be different for $X$ and $Y$. Therefore, we can use a geometric descriptor (LKC) to tell these shapes apart, even though a topological descriptor (Persistence Diagram) fails.

\subsection{Defining the Key Concepts}

\textbf{Intrinsic Volumes ($V_k$):} For 2D sets, these are standard geometric measures:
\begin{itemize}
    \item $V_0$: The Euler Characteristic (for a set in the plane, $V_0 = \#\text{components} - \#\text{holes}$)
    \item $V_1$: Half the boundary length (Perimeter / 2)
    \item $V_2$: The Area
\end{itemize}

\textbf{Lipschitz-Killing Curvatures ($L_k$):} For a compact set with smooth boundary in $\mathbb{R}^2$, the LKCs are essentially the same as the intrinsic volumes, just with a conventional scaling factor for $L_1$:
\begin{align*}
L_0(X_t) &= V_0(X_t) \quad \text{(Euler Characteristic)} \\
L_1(X_t) &= \tfrac{1}{2}V_1(X_t) \quad \text{(Half the Perimeter)} \\
L_2(X_t) &= V_2(X_t) \quad \text{(Area)}
\end{align*}

The ``LKC vector'' $L(X)$ is the collection of these three functions $(L_0(X_t), L_1(X_t), L_2(X_t))$ as the threshold parameter $t$ varies.

\subsection{The Core of the Proof}

\textbf{Assumption of Topological Equivalence:} $X \cong Y$. This directly implies $d_B(\text{PD}(X), \text{PD}(Y)) = 0$ for filtrations based on homeomorphism-invariant functions.

\textbf{Assumption of Geometric Difference at $t^*$:} There exists a specific $t^*$ and an index $k$ such that:
\[
V_k(X_{t^*}) \neq V_k(Y_{t^*})
\]

\textbf{Relating $V_k$ to $L_k$:} By the definitions above:
\begin{itemize}
    \item If $k=0$: $L_0(X_{t^*}) = V_0(X_{t^*}) \neq V_0(Y_{t^*}) = L_0(Y_{t^*})$
    \item If $k=1$: $L_1(X_{t^*}) = \tfrac{1}{2}V_1(X_{t^*}) \neq \tfrac{1}{2}V_1(Y_{t^*}) = L_1(Y_{t^*})$
    \item If $k=2$: $L_2(X_{t^*}) = V_2(X_{t^*}) \neq V_2(Y_{t^*}) = L_2(Y_{t^*})$
\end{itemize}

Therefore, for this same $t^*$ and the corresponding LKC component $L_k$, we have:
\[
L_k(X_{t^*}) \neq L_k(Y_{t^*})
\]

\textbf{Conclusion for the LKC Vectors:} The LKC vector $L(X)$ is a function from the parameter space $T$ (which contains $t^*$) to $\mathbb{R}^3$. Two functions are defined to be equal if their values are equal for all inputs in their domain. We have just shown that at the input $t = t^*$, the outputs differ:
\[
L(X)(t^*) \neq L(Y)(t^*)
\]
Consequently, the functions (or ``curve vectors'') themselves are not equal:
\[
L(X) \neq L(Y)
\]

\subsection{Constructing the Functional $\Phi$}

The final part of the lemma is a straightforward consequence of $L(X) \neq L(Y)$. If two objects are different, there exists a way to measure that difference.

The lemma suggests using the $L^2$-distance between the normalised LKC curves. This is a standard and effective choice.

Let $\tilde{L}(X)$ and $\tilde{L}(Y)$ be the normalised versions of $L(X)$ and $L(Y)$ (e.g., normalised by their total variation or max value to ensure scale invariance).

We can then define the functional $\Phi$ as:
\begin{align}
\Phi(L(X), L(Y)) = \|\tilde{L}(X) - \tilde{L}(Y)\|_{L^2(T)} = \\
\left(\int_T \|\tilde{L}(X)(t) - \tilde{L}(Y)(t)\|^2 \, dt\right)^{1/2}
\end{align}
where $\|\cdot\|$ inside the integral is the Euclidean norm in $\mathbb{R}^3$.

\textbf{Why is $\Phi > 0$?} Since $L(X) \neq L(Y)$, their normalised versions are also not equal. The $L^2$-norm of a continuous function is zero if and only if the function is identically zero. The difference function $\tilde{L}(X) - \tilde{L}(Y)$ is not identically zero (we know it's non-zero at $t^*$), therefore its $L^2$-norm must be positive:
\[
\Phi(L(X), L(Y)) > 0
\]

\textbf{The Contrast:} This stands in stark contrast to the topological descriptor:
\[
d_B(\text{PD}(X), \text{PD}(Y)) = 0 \qedhere
\]
\end{proof}

\end{appendix}

\end{document}